\documentclass{article}

\PassOptionsToPackage{hyperfootnotes=false}{hyperref}
\usepackage[final]{corl_2026}
\makeatletter
\renewcommand{\@conferencelocation}{Austin, Texas, USA}
\makeatother
\usepackage[table]{xcolor}
\definecolor{mydarkblue}{rgb}{0,0.08,0.45}
\definecolor{PredictionRow}{RGB}{248,244,240}
\definecolor{OursRow}{RGB}{238,229,221}
\usepackage{amsmath}
\usepackage{graphicx}
\usepackage{wrapfig}
\usepackage{amssymb}
\usepackage{booktabs}
\usepackage{tabularx}
\usepackage{microtype}
\usepackage{multirow}
\usepackage{subcaption}
\usepackage{float}
\usepackage{flafter}

\newcommand{\best}[1]{\textbf{#1}}
\newcommand{\second}[1]{\underline{#1}}
\newsavebox{\plawablationbox}
\newsavebox{\plawablationimg}
\newlength{\plawablationh}
\newlength{\plawablationw}

\title{PLaW-VLA: Predictive Latent World Modeling \\ for Vision-Language-Action Policies}
\hypersetup{
  pdftitle={PLaW-VLA: Predictive Latent World Modeling for Vision-Language-Action Policies},
  pdfsubject={10th Conference on Robot Learning (CoRL 2026)},
  pdfauthor={Yu Liu, Hetian Guo, Tianlv Huang, Ziyi Cai, Wudi Chen, Hantang Wang, Qiutong Liu, Yingzhi Peng, Wei Han, Peijun Tang, Jianan Wang, Zipei Fan, Zhiyuan Zha, Xuan Song}
}
\newcommand{\authorsep}{\hspace{14pt}}
\author{%
  \normalsize\bfseries
  Yu Liu$^{1,2,}$\thanks{Equal contribution.\quad
    $^{\dagger}$Project leads.\quad
    $^{\ddagger}$Corresponding author.}\phantom{$^{*}$}\authorsep
  Hetian Guo$^{1,*}$\authorsep
  Tianlv Huang$^{1}$\authorsep
  Ziyi Cai$^{3}$\\[2pt]
  \normalsize\bfseries
  Wudi Chen$^{1}$\authorsep
  Hantang Wang$^{4}$\authorsep
  Qiutong Liu$^{4}$\authorsep
  Yingzhi Peng$^{5}$\authorsep
  Wei Han$^{1}$\\[2pt]
  \normalsize\bfseries
  Peijun Tang$^{2,\dagger}$\authorsep
  Jianan Wang$^{2,\dagger}$\authorsep
  Zipei Fan$^{1,\ddagger}$\authorsep
  Zhiyuan Zha$^{1}$\authorsep
  Xuan Song$^{1}$\\[6pt]
  \normalsize\normalfont
  $^{1}$Jilin University\authorsep
  $^{2}$Astribot\authorsep
  $^{3}$Harbin Institute of Technology, Shenzhen\\[1pt]
  \normalfont
  $^{4}$The Hong Kong Polytechnic University\authorsep
  $^{5}$The University of Tokyo\\[6.5pt]
  \normalfont
  \href{https://rainyrobo.github.io/PLaW-VLA/}{\textcolor{mydarkblue}{https://rainyrobo.github.io/PLaW-VLA}}
}

\begin{document}
\maketitle
\begin{figure}[H]
  \vspace{-15pt}
  \centering
  \includegraphics[width=\linewidth]{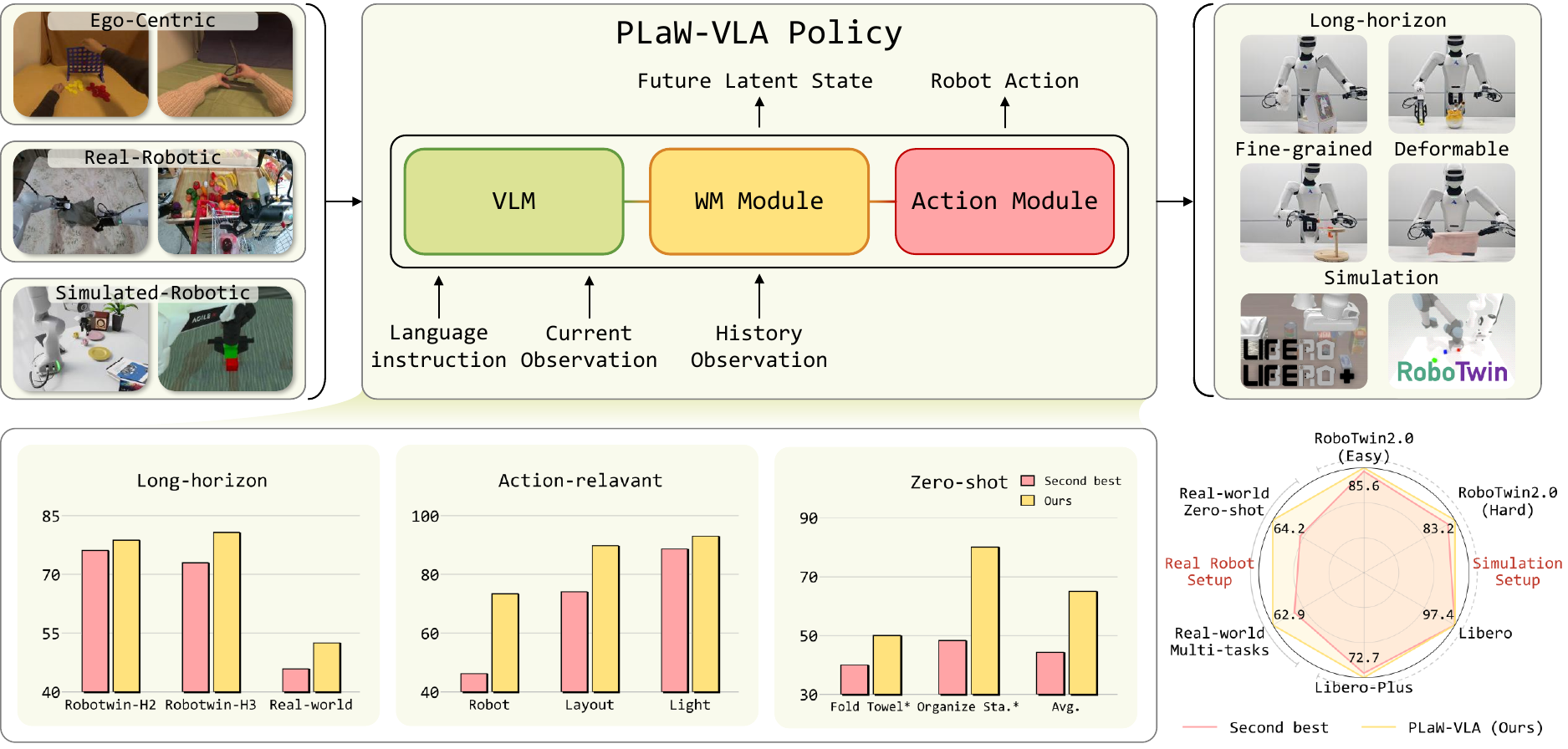}
  \caption{\textbf{PLaW-VLA.} Videos and robot trajectories train a latent world model whose predicted future representations condition robot actions. We evaluate this approach on simulated and real-world manipulation, including multi-step and deformable-object tasks.}
  \label{fig:teaser}
\end{figure}

\begin{abstract}
  Learning to predict how the world evolves can provide vision-language-action (VLA) policies with predictive context for long-horizon control, but its effectiveness depends on what future representation is modeled and how it conditions action generation. We introduce PLaW-VLA, which models task-relevant future states in a pretrained prediction-oriented representation space, reducing the need to predict control-irrelevant visual details. Built on a Mixture-of-Transformers architecture, PLaW-VLA conditions action generation on observation history, current task semantics, and predicted future states through structured causal attention.  Experiments show a $+11.8$ percentage-point (pp) gain over reactive policies on RoboTwin Hard Horizon~III and a $+1.77$ pp gain over reconstruction-oriented latent prediction on zero-shot LIBERO-Plus, supporting improved long-horizon control and generalization under distribution shift, respectively. By avoiding low-level visual reconstruction, PLaW-VLA lowers the burden of future prediction, enabling a lightweight latent world model with parallel future prediction and about $1/19$ the inference latency of generative world--action modeling at comparable policy performance.
\end{abstract}


\section{Introduction}

Vision-language-action (VLA) models adapt pretrained vision-language representations to language-conditioned robot manipulation~\cite{black2025pi_05,black2024pi_0,kim2024openvla,rt2,bjorck2025gr00t,kim2025fine}. Long-horizon manipulation requires not only interpreting the current scene but also tracking how the task evolves across successive interactions. Under partial observability and execution perturbations, the current observation alone may be insufficient to infer the underlying task state. Effective long-horizon control therefore benefits from both modeling temporal context from observation history and anticipating future state evolution for action generation.

Recent studies incorporate predictive representations and future modeling into VLA through future-representation alignment~\citep{zheng2025flare,sun2026vla}, visual feature fusion~\citep{miao2026jepa}, future-conditioned action generation~\cite{cai2026internvla,lv2025f1,pai2025mimic}, and joint world--action modeling~\citep{zhu2025unified,shen2026videovla,ye2026world,bi2025motus,cen2025rynnvla,cen2025worldvla}. This raises two key questions for control: \emph{what future representation should be predicted, and how should it condition action generation?} Existing predictive targets derived from visual reconstruction or generation require modeling both task dynamics and appearance variations, even when the latter are irrelevant to control. This increases the prediction burden, while errors in these irrelevant components may propagate to action generation when the predicted future directly conditions the policy.

We propose PLaW-VLA (Fig.~\ref{fig:teaser}), a VLA framework that integrates predictive latent world modeling with action generation. Built on a Mixture-of-Transformers architecture, PLaW-VLA combines observation history, current task semantics, and predicted future latent states to generate continuous robot actions. The framework is designed around two key choices in future modeling for control. First, PLaW-VLA predicts task-relevant future latent states in a pretrained prediction-oriented representation space, emphasizing state transitions and interaction dynamics rather than low-level visual reconstruction. Second, these predicted future states directly condition the action model through structured causal attention, providing explicit predictive context for action generation.

PLaW-VLA follows a three-stage video-to-robot training pipeline. Stage~I pretrains the latent world model on human and robot videos without action supervision, learning to predict visual dynamics in the representation space defined by the frozen encoder. Stage~II transfers these representations to robot control by introducing action-annotated robot trajectories and jointly optimizing future-state prediction and action generation, thereby aligning predictive dynamics with action-relevant behavior. Stage~III further adapts the resulting policy to the target task and robot embodiment. This training paradigm enables predictive knowledge acquired from large-scale videos to be reused for robot policy learning, providing a scalable path from generic visual dynamics to task-specific control.

We evaluate PLaW-VLA on three simulation benchmarks and diverse real-world manipulation tasks. It outperforms $\pi_{0.5}$ by 11.8 percentage points on RoboTwin Hard Horizon~III and achieves 72.7\% task-weighted success on zero-shot LIBERO-Plus, supporting long-horizon control and generalization under distribution shift. By modeling future dynamics in latent space rather than reconstructing low-level visual content, PLaW-VLA reduces the complexity of future prediction and enables a compact latent world model with parallel prediction, providing an effective trade-off between policy performance and inference efficiency.

\section{Related Work}
\paragraph{Vision-Language-Action Models.}
VLA models transfer pretrained visual and language representations to robot policies~\cite{rt2,team2024octo,kim2024openvla,black2024pi_0,black2025pi_05,bjorck2025gr00t}. Continuous action prediction~\cite{kim2025fine,liang2025discrete}, flow-based decoding~\cite{black2024pi_0,black2025pi_05}, and action tokenization~\cite{pertsch2025fast} provide different interfaces between multimodal representations and robot actions. Complementary approaches augment policy context with memory~\cite{shi2025memoryvla}, spatial representations~\cite{qu2025spatialvla}, and hierarchical task reasoning~\cite{zhao2025cot,huang2026thinkact}. PLaW-VLA complements these directions by augmenting the policy with predicted future visual representations as additional context for action generation.

\paragraph{World Models for Robot Control.}
World models have been increasingly incorporated into robot control by predicting future observations, latent states, or joint video-action trajectories. Visual-imagination approaches generate future images or videos and recover actions through inverse dynamics~\cite{unipi_du2023learning,ko2024learning,tian2025predictive,zhou2024robodreamer}. More recent methods jointly model future observations and actions through autoregressive formulations~\cite{cen2025worldvla,cen2025rynnvla,li2026causal} or video-action generative architectures~\cite{zhu2025unified,shen2026videovla,cai2026internvla,lv2025f1,ye2026world}. Motus~\cite{bi2025motus} unifies understanding, video generation, and action prediction within a Mixture-of-Transformers architecture, while mimic-video~\cite{pai2025mimic} conditions action generation on partially denoised video latents. These methods demonstrate the utility of future prediction for control, but most predictive targets are derived from observation reconstruction or generation objectives. In contrast, PLaW-VLA predicts future features in the space defined by a frozen V-JEPA~2 encoder~\cite{assran2025v}, using a prediction-trained representation space rather than a reconstruction-oriented target space.

\paragraph{Predictive Representations for Policy Learning.}
Predictive representation learning provides an alternative to explicit observation generation by learning latent features that capture temporal structure. FLARE~\cite{zheng2025flare} aligns diffusion-policy features with embeddings of future observations, while VLA-JEPA~\cite{sun2026vla} learns latent action representations through future V-JEPA 2 feature prediction. JEPA-VLA~\cite{miao2026jepa} incorporates V-JEPA 2 visual representations into existing VLAs through feature fusion. These methods differ primarily in how predictive representations interact with policy learning: as auxiliary supervision, latent action representations, or visual features. PLaW-VLA instead predicts future representations from observation history and uses the predicted future state directly to condition action generation. It therefore combines a prediction-trained target space with explicit future-state conditioning for policy learning.

\section{Method}
\label{sec:method}
\begin{figure}[t]
  \centering
  \includegraphics[width=1.0\linewidth]{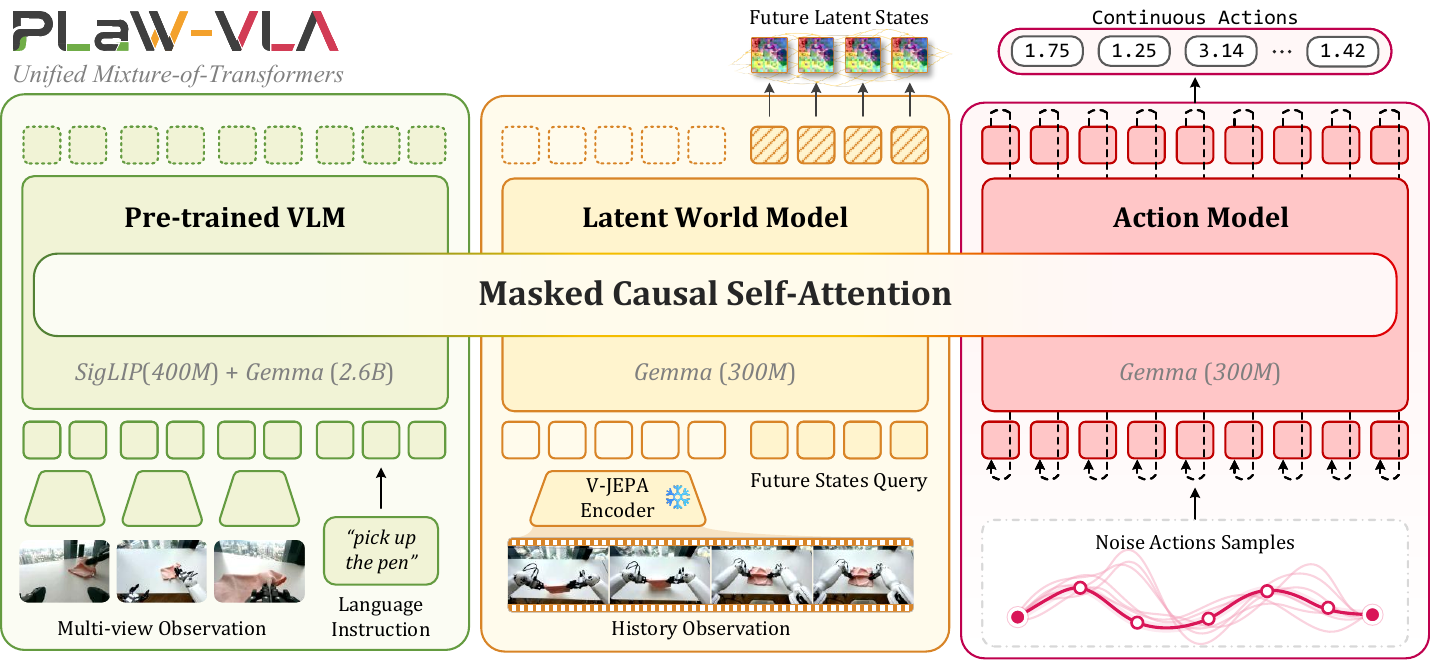}
  \caption{\textbf{PLaW-VLA overview.} Three experts share structured attention for semantic grounding, latent future prediction, and continuous action generation. The world model predicts future representations in a frozen V-JEPA 2 latent space using learnable queries. The action expert attends to the resulting future prediction tokens together with the observed context.}
  \label{fig:framework}
\end{figure}

\subsection{Problem Formulation}

In this work, we formulate vision-language-action generation through predictive latent world modeling. Given a language instruction $c$ and visual observation history $o_{\leq t}$, the model first grounds the current scene into a semantic representation $\phi_t$, predicts future latent states $\hat{\mathbf{z}}_{>t}$, and finally generates future actions $\mathbf{a}_{\geq t}$. This formulation decomposes action generation into three causally ordered components, namely semantic grounding, latent-space prediction, and action decoding. Formally, we factorize the joint conditional distribution over scene semantics $\phi_t$, future latent states $\hat{\mathbf{z}}_{>t}$, and future actions $\mathbf{a}_{\geq t}$ as:
\begin{equation}
  \underbrace{
    p_{\theta}(\mathbf{a}_{\geq t}, \hat{\mathbf{z}}_{>t}, \phi_t \mid o_{\leq t}, c)
  }_{\text{PLaW-VLA}}
  =
  \underbrace{
    p_{\phi}(\phi_t \mid o_t, c)
  }_{\text{VLM}}
  \underbrace{
    p_{z}(\hat{\mathbf{z}}_{>t} \mid o_{\leq t}, \phi_t)
  }_{\text{LWM}}
  \underbrace{
    p_{a}(\mathbf{a}_{\geq t} \mid \phi_t, \mathbf{z}_{t-H_o:t}, \hat{\mathbf{z}}_{>t})
  }_{\text{Action Model}}.
  \label{eq:plaw_factorization}
\end{equation}
The vision-language model (VLM) $p_{\phi}$ grounds the language instruction in the current multimodal context and produces a task-relevant semantic representation $\phi_t$. Conditioned on this grounded representation and the observation history, the latent world model (LWM) $p_{z}$ predicts future latents $\hat{\mathbf{z}}_{>t}$ that summarize the anticipated evolution of the task-relevant world state. Finally, $p_{a}$ decodes future actions from the grounded semantics, historical latents, and predicted future latents.

\subsection{Model Architecture}

As illustrated in Fig.~\ref{fig:framework}, PLaW-VLA adopts a unified Mixture-of-Transformers (MoT) architecture consisting of three causally ordered experts corresponding to the factorization in Eq.~\eqref{eq:plaw_factorization}, including a vision-language expert for semantic grounding, a latent world model expert for predictive latent modeling, and an action expert for continuous action generation. Interactions among experts are regulated through structured causal attention that preserves the intended causal information flow.

\paragraph{Vision-Language Model.}
We instantiate the vision-language expert from PaliGemma~\citep{beyer2024paligemma}, which provides strong multimodal understanding and language-conditioned visual grounding capabilities. The multi-view observation \(o_t\) and language instruction \(c\) are processed by the VLM expert to produce contextual embeddings \(\phi_t\), which provide grounded semantic context for downstream predictive latent modeling and action generation.

\paragraph{Latent World Model.}
Recent large-scale video generation models have demonstrated strong performance by generating future images or videos~\citep{lv2025f1,cai2026internvla,cen2025rynnvla,ye2026world}. In contrast, our approach performs future prediction in a compact latent representation space. The latent world model predicts future latent trajectories conditioned on historical observations and task context, without explicitly modeling low-level visual appearance.

The observation history \(o_{t-H_o:t}\) is first mapped into a compact visual representation space using a frozen pretrained encoder \(E\), producing latent states \(z_{t-H_o:t}\). To perform future-state prediction, we introduce \(K\) learnable future queries \(\mathbf{s}_{1:K}\), where each query corresponds to one future prediction step. These future queries attend to both the historical latent trajectory and the contextual embeddings \(\phi_t\) produced by the VLM, and jointly predict the corresponding future latent states through the latent world model predictor \(F_z\):
\begin{equation}
  z_{t-H_o:t}=E(o_{t-H_o:t}), \qquad
  \hat{\mathbf{z}}_{t+1:t+K}
  =
  F_z\!\left(
  \mathbf{s}_{1:K}, z_{t-H_o:t}, \phi_t
  \right).
\end{equation}

\begin{wrapfigure}{r}{0.35\linewidth}
  \centering
  \includegraphics[width=0.90\linewidth]{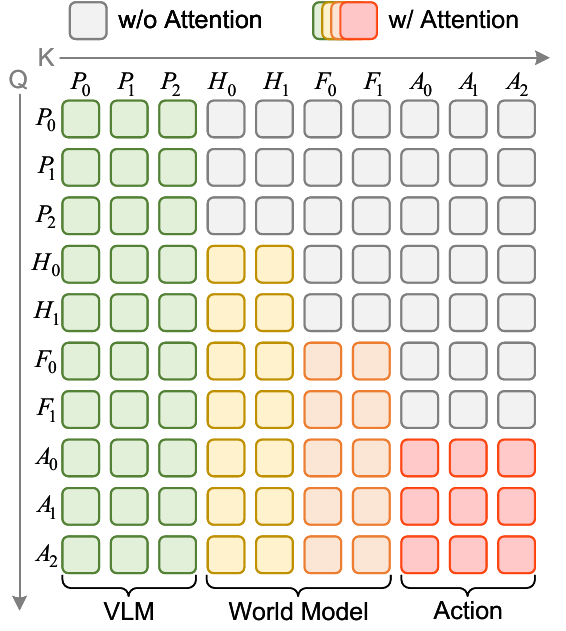}
  \caption{Cross-expert attention.}
  \label{fig:causal_attention}
  \vspace{-3em}
\end{wrapfigure}

\paragraph{Action Model.}
The action expert models the policy \(p_a(\mathbf{a}_{\geq t}\mid \phi_t, \mathbf{z}_{t-H_o:t}, \hat{\mathbf{z}}_{>t})\) by decoding temporally extended action sequences conditioned on grounded semantic representations \(\phi_t\), historical latents \(\mathbf{z}_{t-H_o:t}\), and predicted future latents \(\hat{\mathbf{z}}_{>t}\). The predicted future latents provide predictive context that guides action generation over future interaction horizons.

\paragraph{Causal Cross-Expert Attention.}
As shown in Fig.~\ref{fig:causal_attention}, PLaW-VLA adopts a structured causal attention pattern across the experts. The tokens \(P_i\), \(H_i\), \(F_i\), and \(A_i\) denote contextual, historical latent, future-query latent, and action-query tokens, respectively. This structured attention preserves information flow from semantic grounding to latent prediction and action decoding while preventing future information leakage and reducing cross-expert interference.

\subsection{Training Objective}

We use two complementary objectives: future-state prediction in a pretrained visual representation space and continuous action generation conditioned on historical and predicted future latents.

\paragraph{Latent Predictive Objective.}
The LWM is trained by aligning predicted future states with latent targets extracted from future observations using a frozen pretrained visual encoder, instantiated as V-JEPA 2~\cite{assran2025v}. Its representations provide a compact predictive latent space for future-state supervision, with stop-gradient applied to the target latents:
\begin{equation}
  \mathcal{L}_{\mathrm{wm}}(\mathcal{D})
  =
  \mathbb{E}_{(o_{t-H_o:t+K},c)\sim\mathcal{D}}
  \left[
  \frac{1}{K}
  \sum_{i=1}^{K}
  \left\|
  \hat{z}_{t+i}
  -
  \mathrm{sg}
  \left(
  z_{t+i}
  \right)
  \right\|^2
  \right],
\end{equation}

\paragraph{Action Generation Objective.}
Following $\pi_0$~\citep{black2024pi_0}, the action expert is optimized with a flow-matching objective over continuous action chunks. Let $\mathbf{a}^{\tau}_{t:t+H_a-1}=\tau\boldsymbol{\epsilon}+(1-\tau)\mathbf{a}_{t:t+H_a-1}$ denote the interpolated action sample, where $\boldsymbol{\epsilon}$ is standard Gaussian noise with the same shape as the action chunk and $\tau\sim\mathrm{Beta}(1.5,1)$. The action-generation loss is therefore represented as:

\begin{equation}
  \mathcal{L}_{\mathrm{FM}}(\mathcal{D})
  =
  \mathbb{E}_{\mathcal{D},\,\tau,\,\boldsymbol{\epsilon}}
  \left[
  \left\|
  \pi_{\theta}
  \left(
  \phi_t,
  \mathbf{z}_{t-H_o:t},
  \hat{\mathbf{z}}_{t+1:t+K},
  \mathbf{a}^{\tau}_{t:t+H_a-1},
  \tau
  \right)
  -
  \left(
  \boldsymbol{\epsilon}
  -
  \mathbf{a}_{t:t+H_a-1}
  \right)
  \right\|^2
  \right].
\end{equation}

\paragraph{Joint Robot Policy Learning.}
During robot policy learning, we jointly optimize future-state prediction and action generation. To avoid over-reliance on predictive context, we stochastically drop the predictive branch for each batch. Specifically, we sample $m\sim\mathrm{Bernoulli}(1-\alpha)$ and optimize
\begin{equation}
  \mathcal{L}_{\mathrm{joint}}
  =
  \mathcal{L}_{\mathrm{FM}}(\mathcal{D};\mathcal{C}_m)
  +
  m\lambda_{\mathrm{wm}}\mathcal{L}_{\mathrm{wm}}(\mathcal{D}),
  \qquad
  \mathcal{C}_m=
  \begin{cases}
    (\mathbf{z}_{t-H_o:t},\hat{\mathbf{z}}_{t+1:t+K}), & m=1,\\
    \varnothing, & m=0.
  \end{cases}
  \label{eq:joint_training_loss}
\end{equation}
When $m=1$, the action expert uses historical and predicted future latents; when $m=0$, both groups are omitted from action conditioning and $\mathcal{L}_{\mathrm{wm}}$ is disabled, while the action-generation objective remains active. Training follows three stages: Stage~I pretrains the latent world model on action-free videos, Stage~II jointly learns prediction and action generation on robot trajectories, and Stage~III applies the same objective for task- and embodiment-specific adaptation. Dataset and training details are provided in Appendices~\ref{sec:appendix_data_corpus} and~\ref{sec:appendix_training_details}. Sensitivity to the dropout rate $\alpha$ is reported in Appendix~\ref{app:hyperparameter}.

\section{Experiments}

\subsection{Simulation Benchmarks}

\begin{table}[t]
  \centering
  \caption{\textbf{LIBERO results.} Task success rates (\%). The best and second-best suite scores are indicated by \textbf{boldface} and \underline{underlining}, respectively. \colorbox{PredictionRow}{Shading} marks methods with explicit future-observation prediction.}
  \label{tab:libero}
  \small
  \setlength{\tabcolsep}{4.5pt}
  \begin{tabularx}{\linewidth}{>{\raggedright\arraybackslash}Xccccc}
    \toprule
    Method & LIBERO-Spatial & LIBERO-Object & LIBERO-Goal & LIBERO-Long & Average \\
    \midrule
    $\pi_0$~\cite{black2024pi_0}      & 96.8 & 98.8 & 95.8 & 85.2 & 94.2 \\
    $\pi_{0.5}$~\cite{black2025pi_05} & \second{98.8} & 98.2 & \best{98.0} & 92.4 & 96.9 \\
    FLOWER~\cite{reuss2025flower}     & 97.5 & \second{99.1} & 96.1 & 94.9 & 96.9 \\
    JALA-dino~\cite{luo2026joint}     & 96.0 & 98.2 & 97.4 & \best{96.0} & 96.9 \\
    OpenVLA-OFT~\cite{kim2025fine}    & 97.6 & 98.4 & \second{97.9} & 94.5 & 97.1 \\
    \rowcolor{PredictionRow} UniVLA~\cite{wang2025unified} & 95.4 & 98.8 & 93.6 & 94.0 & 95.5 \\
    \rowcolor{PredictionRow} F1~\cite{lv2025f1} & 98.2 & 97.8 & 95.4 & 91.3 & 95.7 \\
    \rowcolor{PredictionRow} VLA-JEPA~\cite{sun2026vla} & 96.2 & \best{99.6} & 97.2 & \second{95.8} & 97.2 \\
    \rowcolor{OursRow} Ours & $\mathbf{98.9} \pm 0.6$ & $99.0 \pm 0.3$ & $97.1 \pm 1.0$ & $94.6 \pm 0.8$ & \best{97.4} \\
    \bottomrule
  \end{tabularx}
\end{table}

\begin{table}[t]
  \centering
  \caption{\textbf{LIBERO-Plus results.} Zero-shot success rates (\%); averages use official task weights (UniVLA recomputed). \textbf{Boldface} and \underline{underlining} indicate best and second-best scores. \colorbox{PredictionRow}{Shading} marks explicit future-observation prediction.}
  \label{tab:libero_plus}
  \small
  \setlength{\tabcolsep}{5pt}
  \begin{tabularx}{\linewidth}{>{\raggedright\arraybackslash}Xcccccccc}
    \toprule
    Method & Robot & Layout & Light & Background & Language & Noise & Camera & Average \\
    \midrule
    $\pi_0$~\cite{black2024pi_0}                      & 6.0  & 68.9 & 85.0 & 81.4 & 58.8 & \best{79.0} & 13.8 & 53.6 \\
    $\pi_0$-Fast~\cite{black2024pi_0,pertsch2025fast} & 21.6 & 68.8 & 73.2 & 73.2 & 61.0 & 74.4 & \best{65.1} & 61.6 \\
    RIPT-VLA~\cite{tan2025interactive}                & 31.2 & \second{74.2} & 88.4 & 91.6 & \second{77.6} & 73.5 & 55.2 & 68.4 \\
    OpenVLA-OFT~\cite{kim2025fine}                    & 31.9 & \second{74.2} & \second{88.7} & \second{93.3} & \best{79.5} & \second{75.8} & \second{56.4} & 69.6 \\
    \rowcolor{PredictionRow} WorldVLA~\cite{cen2025worldvla} & 27.9 & 38.0 & 43.7 & 17.1 & 41.6 & 10.9 & 0.1 & 25.0 \\
    \rowcolor{PredictionRow}
    UniVLA~\cite{bu2025univla} & \second{46.2} & 31.9 & 69.0 & 81.0 & 69.6 & 21.2 & 1.8 & 42.9 \\
    \rowcolor{OursRow} Ours & \best{73.5} & \best{89.9} & \best{93.1} & \best{93.6} & 75.7 & 55.2 & 41.5 & \best{72.7} \\
    \bottomrule
  \end{tabularx}
\end{table}

\paragraph{LIBERO.}
Following $\pi_0$~\cite{black2024pi_0}, we fine-tune the policy on LIBERO~\cite{liu2023libero} using 1,693 demonstrations spanning LIBERO-Spatial, Object, Goal, and Long. Each suite contains 10 tasks, with 50 evaluation rollouts per task and results averaged over three evaluation seeds. For each suite, $\pm$ denotes the standard deviation across three evaluation seeds. Table~\ref{tab:libero} compares representative VLA policies; the full comparison appears in Appendix~\ref{sec:appendix_libero_results}.

PLaW-VLA achieves the highest average success rate among the compared methods, reaching 97.4\%, which is 0.5 percentage points above its $\pi_{0.5}$ backbone. The improvement is most pronounced on LIBERO-Long, where success increases from 92.4\% to 94.6\%, indicating a clearer benefit on long-horizon tasks. This pattern aligns with the design of PLaW-VLA: the predictive world model leverages observation history to anticipate future states, providing the policy with richer temporal context for tasks that require sustained state tracking and multi-stage execution.

\paragraph{LIBERO-Plus.}
We evaluate the LIBERO-adapted policy zero-shot on the seven perturbation types of LIBERO-Plus.
The Average column uses official task counts, and baseline results come from the official leaderboard~\cite{fei2025libero}. The full comparison appears in Appendix~\ref{sec:appendix_libero_plus_results}.

Table~\ref{tab:libero_plus} shows that PLaW-VLA achieves a task-weighted average success rate of 72.7\%, surpassing the zero-shot baselines from the official LIBERO-Plus leaderboard included in this comparison. Among these methods, it also performs best under Robot, Layout, Light, and Background perturbations, which alter robot state, spatial layout, or scene appearance while largely preserving task-relevant objects, interactions, and goals. This supports the use of a prediction-oriented latent space for modeling task-relevant state evolution rather than low-level visual details. Noise and Camera perturbations more directly disrupt the visual observations and thus affect future-state prediction. Even under these shifts, PLaW-VLA outperforms WorldVLA across all seven perturbation types, indicating that latent prediction reduces the influence of visual detail on future-state modeling and limits its effect on action generation.

\begin{table}[t]
  \centering
  \caption{\textbf{Results on RoboTwin.} Success rates (\%) are reported. The best results are shown in \textbf{bold}, the second-best results are \underline{underlined}, and values in parentheses indicate the absolute gains of PLaW-VLA over the second-best method.}
  \label{tab:robotwin}
  \small
  \setlength{\tabcolsep}{4.5pt}
  \begin{tabular*}{\linewidth}{l@{\extracolsep{\fill}}cccccccc}
    \toprule
    \multirow{2}{*}{Metric}
    & \multicolumn{2}{c}{X-VLA~\cite{zheng2025x}}
    & \multicolumn{2}{c}{$\pi_0$~\cite{black2024pi_0}}
    & \multicolumn{2}{c}{$\pi_{0.5}$~\cite{black2025pi_05}}
    & \multicolumn{2}{c}{PLaW-VLA (Ours)} \\
    \cmidrule(lr){2-3}
    \cmidrule(lr){4-5}
    \cmidrule(lr){6-7}
    \cmidrule(lr){8-9}
    & Easy & Hard
    & Easy & Hard
    & Easy & Hard
    & Easy & Hard \\
    \midrule
    Horizon I
    & 81.6 & \underline{82.5}
    & 66.5 & 61.6
    & \underline{85.1} & 80.2
    & \textbf{87.9}{\scriptsize$_{(+2.8)}$}
    & \textbf{87.8}{\scriptsize$_{(+5.3)}$} \\
    Horizon II
    & 59.3 & 55.9
    & 66.1 & 54.7
    & \underline{79.3} & \underline{73.0}
    & \textbf{82.1}{\scriptsize$_{(+2.8)}$}
    & \textbf{75.5}{\scriptsize$_{(+2.5)}$} \\
    Horizon III
    & 61.2 & 66.0
    & 61.6 & 50.2
    & \underline{78.6} & \underline{67.4}
    & \textbf{82.4}{\scriptsize$_{(+3.8)}$}
    & \textbf{79.2}{\scriptsize$_{(+11.8)}$} \\
    \midrule
    \textbf{Average}
    & 72.9 & 72.8
    & 65.9 & 58.4
    & \underline{82.7} & \underline{76.8}
    & \textbf{85.6}{\scriptsize$_{(+2.9)}$}
    & \textbf{83.2}{\scriptsize$_{(+6.4)}$} \\
    \bottomrule
  \end{tabular*}
\end{table}

\paragraph{RoboTwin.}
Following the evaluation setting adopted in~\cite{li2026causal}, we evaluate PLaW-VLA on 50 RoboTwin~2.0~\cite{chen2025robotwin} tasks. The training data consist of 2,500 demonstrations collected in clean scenes (50 per task) and 25,000 demonstrations collected in randomized scenes (500 per task). We group tasks into three horizons by the number of sequential steps (Appendix~\ref{sec:appendix_robotwin_results}).

As shown in Table~\ref{tab:robotwin}, PLaW-VLA performs best among the compared methods on both Easy and Hard. Relative to $\pi_{0.5}$, average success rises from 82.7\% to 85.6\% on Easy and from 76.8\% to 83.2\% on Hard. The largest gain is on Hard Horizon~III, from 67.4\% to 79.2\% (+11.8 points). This concentration of gains under longer, randomized execution suggests that predictive context helps maintain task progress.

\subsection{Real-World Evaluation}
We evaluate PLaW-VLA on a dual-arm wheeled humanoid robot across four tasks: Fold Towel, Pack Toy, Play Basketball, and Organize Stationery, covering deformable-object manipulation, multi-stage execution, and precise placement. PLaW-VLA, $\pi_0$~\cite{black2024pi_0}, and $\pi_{0.5}$~\cite{black2025pi_05} use the same 400 demonstrations per task and the same task-specific training budget. Each setting includes 10 spatial configurations with three trials per configuration. We report normalized task progress, with detailed scoring rules provided in Appendix~\ref{sec:appendix_task_scoring}.

\begin{wrapfigure}{r}{0.6\linewidth}
  \centering
  \vspace{-2em}
  \includegraphics[width=0.95\linewidth]{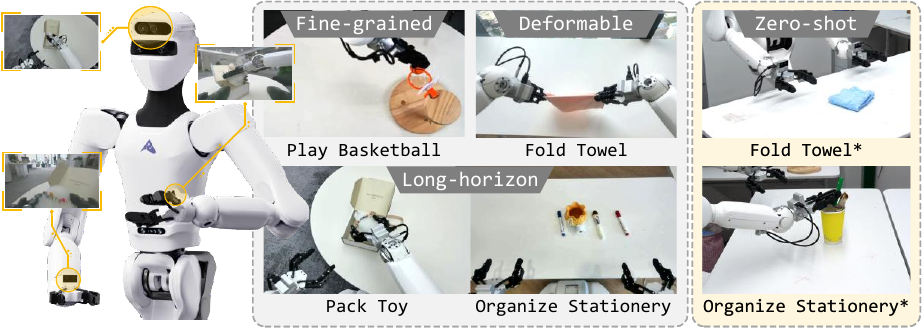}
  \vspace{1em}
  \includegraphics[width=0.99\linewidth]{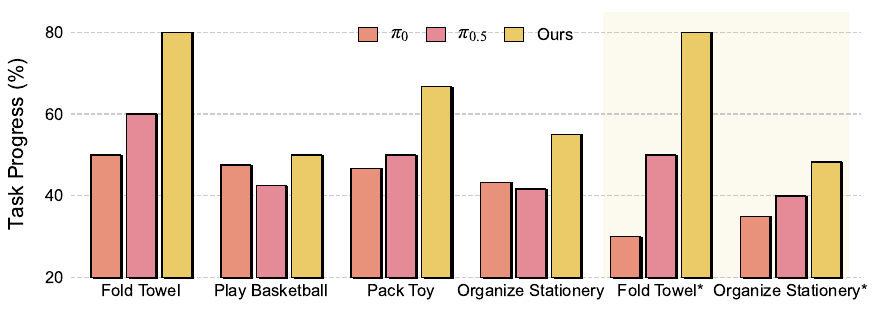}
  \vspace{-0.5em}
  \caption{\textbf{Real-world robot evaluation.} Task progress across four manipulation tasks on a dual-arm wheeled humanoid robot. * denotes zero-shot variants with unseen object appearances or receptacles.}
  \label{fig:real_world_experiment}
  \vspace{-3em}
\end{wrapfigure}

As shown in Fig.~\ref{fig:real_world_experiment}, PLaW-VLA outperforms both baselines on all four seen tasks, with the largest gains on Fold Towel and Pack Toy. The former requires tracking intermediate towel configurations across successive folds, while the latter involves grasping, placement, and box closure. Both tasks therefore require action generation to account for intermediate states as the interaction progresses. This pattern is consistent with the larger gain observed on RoboTwin Hard Horizon III and supports the use of historical and predictive context for maintaining task progress during multi-stage execution.

\paragraph{Zero-shot variants.}
We additionally evaluate two variants without further adaptation. Fold Towel* uses unseen towel materials and colors, while Organize Stationery* replaces the original pen holder with a paper cup that differs in height, opening diameter, and color. PLaW-VLA maintains higher task progress than both baselines in both settings. These results extend the generalization evaluation beyond simulation to changes in object appearance and receptacle geometry while preserving the task goal.

\subsection{Ablation Studies}

Table~\ref{tab:stage_ablation} follows a top-down ablation path from the overall contribution of predictive world modeling to its internal mechanism and inference-time use, structured around the following five questions:

\begin{table}[t]
  \centering
  \caption{\textbf{Predictive world-model contribution and use for control.} Success rates averaged over three evaluation seeds (\%). Parentheses give percentage-point changes relative to PLaW-VLA (full). Group~(a) omits Stage~I and use matched Stage~II data. Masking is inference-only; a dash means not evaluated.}
  \label{tab:stage_ablation}
  \setlength{\plawablationw}{258pt}%
  \sbox{\plawablationbox}{%
    \begin{minipage}{\plawablationw}
      \small
\setlength{\tabcolsep}{3pt}
\renewcommand{\arraystretch}{1}
\newcommand{\abscore}[2]{#1\,{\footnotesize\textcolor[RGB]{124,63,59}{(#2)}}}
\begin{tabularx}{\linewidth}{l>{\centering\arraybackslash}X>{\centering\arraybackslash}X}
  \toprule
  \textbf{Setting} & \textbf{LIBERO} & \textbf{LIBERO-Plus} \\
  \midrule
  \rowcolor[RGB]{244,244,244}
  \textbf{PLaW-VLA (full)} & \textbf{97.40} & \textbf{72.70} \\
  \addlinespace[2pt]
  \multicolumn{3}{@{}l}{\textcolor[RGB]{102,102,102}{\textit{a) World-model contribution (w/o Stage~I)}}} \\
  \quad PLaW-VLA (w/o Stage~I) & \abscore{96.28}{-1.12} & \abscore{69.05}{-3.65} \\
  \quad w/o world model & \abscore{96.05}{-1.35} & \abscore{65.49}{-7.21} \\
  \addlinespace[2pt]
  \multicolumn{3}{@{}l}{\textcolor[RGB]{102,102,102}{\textit{b) Prediction-to-control mechanism}}} \\
  \quad w/o prediction loss & \abscore{96.30}{-1.10} & \abscore{67.56}{-5.14} \\
  \quad w/o future conditioning & \abscore{96.85}{-0.55} & \abscore{69.81}{-2.89} \\
  \addlinespace[2pt]
  \multicolumn{3}{@{}l}{\textcolor[RGB]{102,102,102}{\textit{c) Prediction space}}} \\
  \quad Cosmos VAE & \abscore{96.65}{-0.75} & \abscore{70.93}{-1.77} \\
  \addlinespace[2pt]
  \multicolumn{3}{@{}l}{\textcolor[RGB]{102,102,102}{\textit{d) Predictive pretraining}}} \\
  \quad w/o Stage~I & \abscore{96.28}{-1.12} & \abscore{69.05}{-3.65} \\
  \quad w/o Stage~II & \abscore{95.65}{-1.75} & \abscore{66.03}{-6.67} \\
  \quad w/o Stage~I and II & \abscore{94.05}{-3.35} & \abscore{60.80}{-11.90} \\
  \addlinespace[2pt]
  \multicolumn{3}{@{}l}{\textcolor[RGB]{102,102,102}{\textit{e) Use of predicted future states}}} \\
  \quad w/o all future states & \abscore{93.05}{-4.35} & --- \\
  \quad w/o far-future state & \abscore{95.45}{-1.95} & --- \\
  \quad w/o near-future state & \abscore{95.80}{-1.60} & --- \\
  \bottomrule
\end{tabularx}
    \end{minipage}%
  }%
  \setlength{\plawablationh}{\dimexpr\ht\plawablationbox+\dp\plawablationbox\relax}%
  \sbox{\plawablationimg}{%
    \includegraphics[height=0.99\plawablationh]{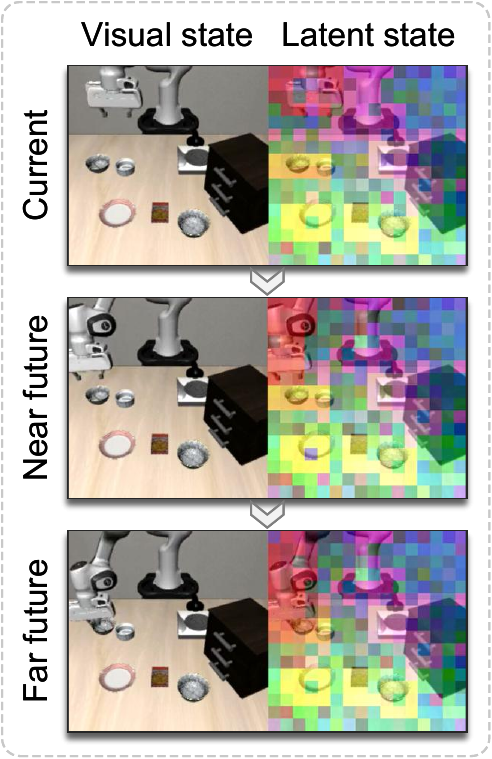}%
  }%
  \noindent
  \raisebox{\dp\plawablationbox}[\plawablationh][0pt]{\usebox{\plawablationbox}}%
  \hfill
  \usebox{\plawablationimg}%
\end{table}

\paragraph{a) Does predictive world modeling itself help?}
We isolate the effect of predictive world modeling under matched training data. Both variants remove Stage I and use identical Stage II robot data and training budgets; the ablated variant further removes predictive world modeling. This lowers success from 96.28\% to 96.05\% on LIBERO and from 69.05\% to 65.49\% on LIBERO-Plus, with a more pronounced 3.56-point drop on LIBERO-Plus. These results show that predictive world modeling provides gains even without additional video pretraining, rather than merely benefiting from greater data exposure.

\paragraph{b) How does predictive world modeling improve the policy?}
We disentangle the roles of predicting future states and conditioning action generation on them. Removing the prediction loss reduces LIBERO-Plus success from 72.70\% to 67.56\% (-5.14 points), whereas retaining predictive supervision but removing future conditioning reduces it to 69.81\% (-2.89 points). The corresponding drops on LIBERO are 1.10 and 0.55 points. This shows that predictive supervision is important for learning informative future representations, while future conditioning allows the action expert to exploit them for control. Their complementary contributions explain the benefit of predictive world modeling.

\paragraph{c) Which representation space should be used for future prediction?}
We replace V-JEPA 2 with Cosmos VAE~\cite{agarwal2025cosmos} while keeping the predictive pipeline unchanged. The VAE variant achieves 96.65\% on LIBERO and 70.93\% on LIBERO-Plus, compared with 97.40\% and 72.70\% for V-JEPA 2. The predictive formulation remains effective with reconstruction-oriented VAE latents, showing that its benefit is not tied to a specific representation space. However, the stronger performance with V-JEPA 2 shows that predictive world modeling depends not only on whether the future is predicted, but also on where it is predicted. A prediction-oriented latent space therefore provides a more effective predictive context for downstream action generation.

\paragraph{d) How is this predictive capability acquired?}
We next examine the roles of the two pretraining stages. Removing Stage~I decreases LIBERO-Plus success by 3.65 percentage points, while removing Stage~II causes a larger 6.67-point drop; removing both leads to an 11.90-point decrease. The corresponding drops on LIBERO are 1.12, 1.75, and 3.35 points, respectively. These results show that both video-based predictive pretraining in Stage~I and robot-policy pretraining in Stage~II contribute to the final performance. Stage~I establishes predictive dynamics from action-free videos, whereas Stage~II further grounds this capability in robot trajectories by jointly learning future-state prediction and action generation.

\paragraph{e) Does the policy use predicted future states?}
At inference, masking all predicted future states reduces LIBERO success from 97.40\% to 93.05\%, while masking only the far- or near-future state reduces success to 95.45\% and 95.80\%, respectively. Both horizons contribute to average success, with suite-level exceptions. The slightly larger drop from masking the far-future state suggests that longer-horizon predictions provide information about later task progression beyond the immediate future. The PCA visualization in Table~\ref{tab:stage_ablation} shows that the predicted features change from the current state to the near- and far-future states together with the robot arm motion, consistent with modeling the evolution of the interaction. Horizon definitions and suite-level results are provided in Appendix~\ref{sec:appendix_future_masking}.

Additional analyses of training dynamics and checkpoint robustness are provided in Appendix~\ref{app:additional_analysis}, and hyperparameter sensitivity in Appendix~\ref{app:hyperparameter}.

\subsection{Inference Efficiency}

\begin{wrapfigure}{r}{0.4\linewidth}
  \centering
  \vspace{-1.5em}
  \includegraphics[width=0.98\linewidth]{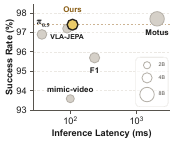}
  \caption{\textbf{Inference efficiency.} LIBERO success versus single-H100 inference latency under matched settings. Marker size indicates model parameter count. The mimic-video score averages three suites.}
  \label{fig:inference_tradeoff}
  \vspace{-0.8em}
\end{wrapfigure}

Fig.~\ref{fig:inference_tradeoff} further evaluates the inference cost of future modeling. Compared with its $\pi_{0.5}$ backbone, PLaW-VLA improves LIBERO success from 96.9\% to 97.4\% while adding only 0.9B parameters and approximately 70\,ms of inference latency. Compared with Motus~\citep{bi2025motus}, PLaW-VLA requires only about $1/19$ of the inference latency at comparable LIBERO success. This difference is consistent with their distinct future-modeling paradigms: Motus adopts a unified world--action generative formulation, whereas PLaW-VLA predicts control-relevant future states in a prediction-oriented latent space. By avoiding explicit modeling of low-level appearance details such as texture and background, PLaW-VLA uses a lightweight latent world model in which multiple future states are predicted in parallel through future queries and directly used to condition the action expert, without additional video generation or decoding. PLaW-VLA therefore introduces predictive future context into action generation with limited additional online inference cost.

\section{Conclusion}
We focus on how future states should be modeled and used in VLA policies and introduce PLaW-VLA, which models task-relevant state evolution in a prediction-oriented latent space and directly conditions action generation on predicted futures. This design increases the control relevance of future representations while reducing the influence of control-irrelevant prediction errors on policy outputs. Experiments show that PLaW-VLA improves the performance--efficiency trade-off, achieving a $+11.8$ pp gain over $\pi_{0.5}$ on RoboTwin Hard Horizon~III, 72.7\% average success on zero-shot LIBERO-Plus, and about $1/19$ the inference latency of generative world--action modeling at comparable policy performance. Ablations further identify the contributions of predictive supervision, future conditioning, the representation space, and pretraining. Together, these findings highlight the importance of both what a VLA policy predicts and how those predictions inform its actions.

\section{Limitations}

PLaW-VLA has two main limitations. First, future prediction and closed-loop control are coupled through execution errors: trajectory deviations can degrade future-state prediction, while inaccurate predictions may in turn affect subsequent actions through future-conditioned control. Addressing this coupling may require predictive uncertainty estimation, adaptive prediction horizons, and recovery mechanisms.
Second, the scalability of the video-to-robot training paradigm remains underexplored. Although Stage~I requires no action annotations and can leverage large-scale action-free videos, the current training data remain limited in scale and diversity. A key next step is to study how video scale and data distribution affect transfer to downstream robot control and whether broader pretraining yields sustained policy gains.

\clearpage
\bibliography{references}
\newpage
\appendix

\section{Additional Benchmark Results}

\subsection{Extended LIBERO Results}
\label{sec:appendix_libero_results}
Table~\ref{tab:libero_full} extends the comparison in Table~\ref{tab:libero} to additional baselines. PLaW-VLA achieves the highest average success among the listed methods, at 97.4\%. The relative ranking nevertheless varies across suites: VLA-JEPA achieves higher success on LIBERO-Goal and LIBERO-Long. The overall advantage therefore coexists with suite-specific differences rather than a uniform improvement across all tasks.
\begin{table}[H]
  \centering
  \caption{Performance comparison on LIBERO. CoT-VLA and MolmoAct averages are recomputed from the displayed suite scores. The mimic-video average covers only the three reported suites.}
  \label{tab:libero_full}
  \resizebox{\linewidth}{!}{
    \begin{tabular}{l ccccc}
      \toprule
      Method & LIBERO-Spatial & LIBERO-Object & LIBERO-Goal & LIBERO-Long & Average \\
      \midrule
      TraceVLA~\cite{zheng2025tracevla}     & 84.6 & 85.2 & 75.1 & 54.1 & 74.8 \\
      Octo~\cite{team2024octo}              & 78.9 & 85.7 & 84.6 & 51.1 & 75.1 \\
      OpenVLA~\cite{kim2024openvla}         & 84.7 & 88.4 & 79.2 & 53.7 & 76.5 \\
      SpatialVLA~\cite{qu2025spatialvla}    & 88.2 & 89.9 & 78.6 & 55.5 & 78.1 \\
      LAPA~\cite{ye2025latent}              & 83.4 & 87.6 & 78.2 & 68.8 & 79.5 \\
      WorldVLA~\cite{cen2025worldvla}       & 87.6 & 96.2 & 83.4 & 60.0 & 81.8 \\
      CoT-VLA~\cite{zhao2025cot}            & 87.5 & 91.6 & 87.6 & 69.0 & 83.9 \\
      ThinkAct~\cite{huang2026thinkact}     & 88.3 & 91.4 & 87.1 & 70.9 & 84.4 \\
      $\pi_0$+FAST~\cite{black2024pi_0,pertsch2025fast}     & 96.4 & 96.8 & 88.6 & 60.2 & 85.5 \\
      MolmoAct~\cite{lee2025molmoact}       & 87.0 & 95.4 & 87.6 & 77.2 & 86.8 \\
      SmolVLA~\cite{shukor2025smolvla}      & 93.0 & 94.0 & 91.0 & 77.0 & 88.8 \\
      Being-H0~\cite{luo2025being}          & 92.6 & 96.8 & 94.0 & 77.4 & 90.2 \\
      GR00T-N1~\cite{bjorck2025gr00t}       & 94.4 & 97.6 & 93.0 & 90.6 & 93.9 \\
      mimic-video~\cite{pai2025mimic}       & 94.2 & 96.8 & 90.6 & --   & 93.9 \\
      $\pi_0$~\cite{black2024pi_0}          & 96.8 & 98.8 & 95.8 & 85.2 & 94.2 \\
      UniVLA~\cite{wang2025unified}         & 95.4 & 98.8 & 93.6 & 94.0 & 95.5 \\
      F1~\cite{lv2025f1}                    & 98.2 & 97.8 & 95.4 & 91.3 & 95.7 \\
      DD-VLA~\cite{liang2025discrete}       & 97.2 & 98.6 & 97.4 & 92.0 & 96.3 \\
      FLOWER~\cite{reuss2025flower}         & 97.5 & 99.1 & 96.1 & 94.9 & 96.9 \\
      $\pi_{0.5}$~\cite{black2025pi_05}     & 98.8 & 98.2 & 98.0 & 92.4 & 96.9 \\
      JALA-dino~\cite{luo2026joint}         & 96.0 & 98.2 & 97.4 & 96.0 & 96.9 \\
      CronusVLA~\cite{li2025cronusvla}      & 97.3 & 99.6 & 96.9 & 94.0 & 97.0 \\
      OpenVLA-OFT~\cite{kim2025fine}        & 97.6 & 98.4 & 97.9 & 94.5 & 97.1 \\
      VLA-JEPA~\cite{sun2026vla}            & 96.2 & 99.6 & 97.2 & 95.8 & 97.2 \\
      \midrule
      Ours                                  & $98.9 \pm 0.6$ & $99.0 \pm 0.3$ & $97.1 \pm 1.0$ & $94.6 \pm 0.8$ & 97.4 \\
      \bottomrule
    \end{tabular}
  }
\end{table}

\subsection{Extended LIBERO-Plus Results}
\label{sec:appendix_libero_plus_results}
Table~\ref{tab:libero_plus_full} gives the full LIBERO-Plus comparison, including the OpenVLA and NORA results omitted from Table~\ref{tab:libero_plus} for space. PLaW-VLA has the highest task-weighted average among these methods.
\begin{table}[H]
  \centering
  \caption{\textbf{Full LIBERO-Plus comparison.} Zero-shot task success rates (\%); the average is weighted by official task counts. UniVLA's average is recomputed from its perturbation scores. The best and second-best perturbation scores are indicated by \textbf{boldface} and \underline{underlining}, respectively.}
  \label{tab:libero_plus_full}
  \small
  \setlength{\tabcolsep}{5pt}
  \begin{tabularx}{\linewidth}{>{\raggedright\arraybackslash}Xcccccccc}
    \toprule
    Method & Robot & Layout & Light & Background & Language & Noise & Camera & Average \\
    \midrule
    OpenVLA~\cite{kim2024openvla}                     & 3.5  & 28.5 & 8.1  & 34.8 & 23.0 & 15.2 & 0.8 & 15.6 \\
    NORA~\cite{hung2025nora}                          & 37.0 & 62.1 & 45.7 & 58.6 & 65.1 & 12.8 & 2.2 & 39.0 \\
    $\pi_0$~\cite{black2024pi_0}                      & 6.0  & 68.9 & 85.0 & 81.4 & 58.8 & \best{79.0} & 13.8 & 53.6 \\
    $\pi_0$-Fast~\cite{black2024pi_0,pertsch2025fast} & 21.6 & 68.8 & 73.2 & 73.2 & 61.0 & 74.4 & \best{65.1} & 61.6 \\
    RIPT-VLA~\cite{tan2025interactive}                & 31.2 & \second{74.2} & 88.4 & 91.6 & \second{77.6} & 73.5 & 55.2 & 68.4 \\
    OpenVLA-OFT~\cite{kim2025fine}                    & 31.9 & \second{74.2} & \second{88.7} & \second{93.3} & \best{79.5} & \second{75.8} & \second{56.4} & 69.6 \\
    WorldVLA~\cite{cen2025worldvla} & 27.9 & 38.0 & 43.7 & 17.1 & 41.6 & 10.9 & 0.1 & 25.0 \\
    UniVLA~\cite{bu2025univla} & \second{46.2} & 31.9 & 69.0 & 81.0 & 69.6 & 21.2 & 1.8 & 42.9 \\
    Ours & \best{73.5} & \best{89.9} & \best{93.1} & \best{93.6} & 75.7 & 55.2 & 41.5 & \best{72.7} \\
    \bottomrule
  \end{tabularx}
\end{table}

\clearpage
\subsection{Task-Level RoboTwin Results}
\label{sec:appendix_robotwin_results}
Table~\ref{tab:robotwin_task_results} resolves the horizon averages in Table~\ref{tab:robotwin} into 50 tasks. Easy uses fixed initial configurations, whereas Hard randomizes object poses and scene layouts. Relative to $\pi_{0.5}$, PLaW-VLA achieves higher success on 36 Hard tasks, ties 4, and declines on 10; all 5 Hard Horizon~III tasks improve. The average gain therefore reflects a broad but nonuniform shift. Handover Mic and Place Dual Shoes illustrate regressions that the horizon means alone do not show.
\begin{table}[H]
  \centering
  \caption{\textbf{Evaluation on RoboTwin 2.0 Simulation: Easy vs Hard (50 tasks).} RoboTwin 2.0 is a challenging bimanual manipulation benchmark requiring coordinated dual-arm control. "Easy" uses fixed initial configurations, while "Hard" involves randomized object poses and scene layouts.}
  \label{tab:robotwin_task_results}
  \resizebox{\textwidth}{!}{
    \begin{tabular}{l c *{8}{c}}
      \toprule
      \multirow{2}{*}{\textbf{Task}}
      & \multirow{2}{*}{\textbf{Horizon}}
      & \multicolumn{2}{c}{\textbf{X-VLA}~\cite{zheng2025x}}
      & \multicolumn{2}{c}{$\boldsymbol{\pi_0}$~\cite{black2024pi_0}}
      & \multicolumn{2}{c}{$\boldsymbol{\pi_{0.5}}$~\cite{black2025pi_05}}
      & \multicolumn{2}{c}{\textbf{Ours}} \\
      \cmidrule(lr){3-4} \cmidrule(lr){5-6} \cmidrule(lr){7-8} \cmidrule(lr){9-10}
      & & Easy & Hard & Easy & Hard & Easy & Hard & Easy & Hard \\
      \midrule
      \textit{Adjust Bottle} & I & 100\% & 99\% & 99\% & 95\% & 100\% & 99\% & 96\% & 99\% \\
      \textit{Beat Block Hammer} & I & 92\% & 88\% & 79\% & 84\% & 96\% & 93\% & 96\% & 98\% \\
      \textit{Blocks Ranking RGB} & III & 83\% & 83\% & 80\% & 63\% & 92\% & 85\% & 95\% & 89\% \\
      \textit{Blocks Ranking Size} & III & 67\% & 74\% & 14\% & 5\% & 49\% & 26\% & 69\% & 53\% \\
      \textit{Click Alarmclock} & I & 99\% & 99\% & 77\% & 68\% & 98\% & 89\% & 95\% & 91\% \\
      \textit{Click Bell} & I & 100\% & 100\% & 71\% & 48\% & 99\% & 66\% & 100\% & 100\% \\
      \textit{Dump Bin Bigbin} & I & 79\% & 77\% & 88\% & 83\% & 92\% & 97\% & 83\% & 87\% \\
      \textit{Grab Roller} & I & 100\% & 100\% & 98\% & 94\% & 100\% & 100\% & 100\% & 100\% \\
      \textit{Handover Block} & II & 73\% & 37\% & 47\% & 31\% & 66\% & 57\% & 83\% & 72\% \\
      \textit{Handover Mic} & II & 0\% & 0\% & 97\% & 97\% & 98\% & 97\% & 86\% & 81\% \\
      \textit{Hanging Mug} & II & 23\% & 27\% & 14\% & 11\% & 18\% & 17\% & 37\% & 31\% \\
      \textit{Lift Pot} & I & 99\% & 100\% & 80\% & 72\% & 96\% & 85\% & 90\% & 89\% \\
      \textit{Move Can Pot} & I & 89\% & 86\% & 68\% & 48\% & 51\% & 55\% & 68\% & 74\% \\
      \textit{Move Pillbottle Pad} & I & 73\% & 71\% & 67\% & 46\% & 84\% & 61\% & 86\% & 82\% \\
      \textit{Move Playingcard Away} & I & 93\% & 98\% & 74\% & 65\% & 96\% & 84\% & 95\% & 91\% \\
      \textit{Move Stapler Pad} & I & 78\% & 73\% & 41\% & 24\% & 56\% & 42\% & 83\% & 74\% \\
      \textit{Open Laptop} & I & 93\% & 100\% & 71\% & 81\% & 90\% & 96\% & 90\% & 91\% \\
      \textit{Open Microwave} & I & 79\% & 71\% & 4\% & 32\% & 34\% & 77\% & 43\% & 86\% \\
      \textit{Pick Diverse Bottles} & II & 58\% & 36\% & 69\% & 31\% & 81\% & 71\% & 89\% & 73\% \\
      \textit{Pick Dual Bottles} & II & 47\% & 36\% & 59\% & 37\% & 93\% & 63\% & 96\% & 82\% \\
      \textit{Place A2B Left} & I & 48\% & 49\% & 43\% & 47\% & 87\% & 82\% & 88\% & 83\% \\
      \textit{Place A2B Right} & I & 36\% & 36\% & 39\% & 34\% & 87\% & 84\% & 94\% & 91\% \\
      \textit{Place Bread Basket} & I & 81\% & 71\% & 62\% & 46\% & 77\% & 64\% & 95\% & 92\% \\
      \textit{Place Bread Skillet} & II & 77\% & 67\% & 66\% & 49\% & 85\% & 66\% & 87\% & 83\% \\
      \textit{Place Burger Fries} & II & 94\% & 94\% & 81\% & 76\% & 94\% & 87\% & 86\% & 89\% \\
      \textit{Place Can Basket} & II & 49\% & 52\% & 55\% & 46\% & 62\% & 62\% & 72\% & 61\% \\
      \textit{Place Cans Plasticbox} & II & 97\% & 98\% & 63\% & 45\% & 94\% & 84\% & 97\% & 91\% \\
      \textit{Place Container Plate} & I & 97\% & 95\% & 97\% & 92\% & 99\% & 95\% & 94\% & 96\% \\
      \textit{Place Dual Shoes} & II & 79\% & 88\% & 59\% & 51\% & 75\% & 75\% & 69\% & 50\% \\
      \textit{Place Empty Cup} & I & 100\% & 98\% & 91\% & 85\% & 100\% & 99\% & 93\% & 97\% \\
      \textit{Place Fan} & I & 80\% & 75\% & 66\% & 71\% & 87\% & 85\% & 96\% & 98\% \\
      \textit{Place Mouse Pad} & I & 70\% & 70\% & 20\% & 20\% & 60\% & 39\% & 73\% & 61\% \\
      \textit{Place Object Basket} & II & 44\% & 39\% & 67\% & 70\% & 80\% & 76\% & 94\% & 87\% \\
      \textit{Place Object Scale} & I & 52\% & 74\% & 57\% & 52\% & 86\% & 80\% & 85\% & 84\% \\
      \textit{Place Object Stand} & I & 86\% & 88\% & 82\% & 68\% & 91\% & 85\% & 95\% & 93\% \\
      \textit{Place Phone Stand} & I & 88\% & 87\% & 49\% & 53\% & 81\% & 81\% & 89\% & 86\% \\
      \textit{Place Shoe} & I & 96\% & 95\% & 76\% & 76\% & 92\% & 93\% & 93\% & 94\% \\
      \textit{Press Stapler} & I & 92\% & 98\% & 44\% & 37\% & 87\% & 83\% & 88\% & 83\% \\
      \textit{Put Bottles Dustbin} & III & 74\% & 77\% & 65\% & 56\% & 84\% & 79\% & 81\% & 83\% \\
      \textit{Put Object Cabinet} & II & 46\% & 48\% & 73\% & 60\% & 80\% & 79\% & 71\% & 75\% \\
      \textit{Rotate QRcode} & I & 34\% & 33\% & 74\% & 70\% & 89\% & 87\% & 94\% & 81\% \\
      \textit{Scan Object} & II & 14\% & 36\% & 55\% & 42\% & 72\% & 65\% & 68\% & 67\% \\
      \textit{Shake Bottle Horizontally} & I & 100\% & 100\% & 98\% & 92\% & 99\% & 99\% & 98\% & 99\% \\
      \textit{Shake Bottle} & I & 99\% & 100\% & 94\% & 91\% & 99\% & 97\% & 97\% & 99\% \\
      \textit{Stack Blocks Three} & III & 6\% & 10\% & 72\% & 52\% & 91\% & 76\% & 93\% & 96\% \\
      \textit{Stack Blocks Two} & II & 92\% & 87\% & 93\% & 79\% & 97\% & 100\% & 99\% & 98\% \\
      \textit{Stack Bowls Three} & III & 76\% & 86\% & 77\% & 75\% & 77\% & 71\% & 74\% & 75\% \\
      \textit{Stack Bowls Two} & II & 96\% & 93\% & 94\% & 95\% & 95\% & 96\% & 98\% & 93\% \\
      \textit{Stamp Seal} & I & 76\% & 82\% & 46\% & 33\% & 79\% & 55\% & 84\% & 71\% \\
      \textit{Turn Switch} & I & 40\% & 61\% & 41\% & 42\% & 62\% & 54\% & 57\% & 63\% \\
      \midrule
      \textbf{Average (\%)} &  & 72.88\% & 72.84\% & 65.92\% & 58.40\% & 82.74\% & 76.76\% & 85.64\% & 83.24\% \\
      \bottomrule
    \end{tabular}
  }
\end{table}

\clearpage
\section{Controlled Analyses}
\label{app:additional_analysis}

\subsection{Pretraining and Training Dynamics}
\label{sec:appendix_pretraining}
Tables~\ref{tab:pretraining_libero} and~\ref{tab:pretraining_libero_plus} compare matched Stage~III checkpoints with and without Stages~I--II. On LIBERO, the full model's average advantage narrows from 3.35 points at 30k steps to 1.00 point at 50k steps; the ranking on LIBERO-Long reverses at 50k. On LIBERO-Plus, the 50k advantage remains 2.83 points overall, although the no-pretraining variant performs better under sensor noise. The observed benefit varies across checkpoints and evaluation shifts; it is not a gain in every setting. The full-model Stage~III trajectory is examined in Appendix~\ref{app:training_dynamics}.
\begin{table}[H]
  \centering
  \caption{\textbf{Effect of Stages~I--II pretraining on LIBERO.} Matched Stage~III checkpoints; all values are success rates (\%).}
  \label{tab:pretraining_libero}
  \resizebox{\linewidth}{!}{
    \begin{tabular}{lcccccc}
      \toprule
      Method & ckpt & LIBERO-Spatial & LIBERO-Object & LIBERO-Goal & LIBERO-Long & Average \\
      \midrule
      Without Stages~I--II & 30k & 97.00 & 98.40 & 91.20 & 89.60 & 94.05 \\
      PLaW-VLA & 30k & 98.90 & 99.00 & 97.10 & 94.60 & 97.40 \\
      $\Delta$ & 30k & +1.90 & +0.60 & +5.90 & +5.00 & +3.35 \\
      \midrule
      Without Stages~I--II & 50k & 97.40 & 98.80 & 94.40 & 93.20 & 95.95 \\
      PLaW-VLA & 50k & 98.90 & 98.90 & 98.30 & 91.70 & 96.95 \\
      $\Delta$ & 50k & +1.50 & +0.10 & +3.90 & -1.50 & +1.00 \\
      \bottomrule
    \end{tabular}
  }
\end{table}

\begin{table}[H]
  \centering
  \caption{\textbf{Effect of Stages~I--II pretraining on LIBERO-Plus at 50k steps.} Success rates (\%); the average is weighted by perturbation task counts.}
  \label{tab:pretraining_libero_plus}
  \resizebox{\linewidth}{!}{
    \begin{tabular}{lccccccccc}
      \toprule
      Method & ckpt & Robot & Layout & Light & Background & Language & Noise & Camera & Average \\
      \midrule
      Without Stages~I--II & 50k & 63.23 & 75.08 & 86.78 & 90.52 & 65.91 & 61.40 & 44.59 & 67.79 \\
      PLaW-VLA & 50k & 68.40 & 84.00 & 90.20 & 91.40 & 75.10 & 53.30 & 45.10 & 70.62 \\
      $\Delta$ & 50k & +5.17 & +8.92 & +3.42 & +0.88 & +9.19 & -8.10 & +0.51 & +2.83 \\
      \bottomrule
    \end{tabular}
  }
\end{table}

\subsection{Training Dynamics and Checkpoint Robustness}
\label{app:training_dynamics}
\begin{figure}[t]
  \centering
  \includegraphics[width=0.48\textwidth]{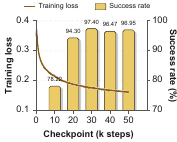}
  \caption{\textbf{Stage~III training dynamics.} Training loss and LIBERO success rate across checkpoints, showing stable downstream performance after 30k updates.}
  \label{fig:training_dynamics}
\end{figure}

To examine whether the Stage~III performance depends on a particular training checkpoint, Fig.~\ref{fig:training_dynamics} reports the training loss and LIBERO success rate throughout Stage~III optimization. The success rate increases rapidly from 78.22\% at 10k updates to 94.30\% at 20k and reaches 97.40\% at 30k. It then remains between 96.47\% and 97.40\% through 50k updates, with less than one percentage point of variation, despite the training loss continuing to decrease. This trajectory indicates a downstream performance plateau after approximately 30k updates, with less than one percentage point of variation across the later evaluated checkpoints. 

\subsection{World-Model Branch Dropout}
\label{app:hyperparameter}
Reducing $\alpha$ from 0.3 to 0.1 gives the same 30k-step LIBERO success rate of 97.4\% and increases training time by approximately 10\% in this setting. We therefore use $\alpha=0.3$. In Eq.~\eqref{eq:joint_training_loss}, $m=0$ omits the world-model branch, including its historical and future tokens, together with $\mathcal{L}_{\mathrm{wm}}$. This stochastic skip is distinct from removing future conditioning.

\subsection{Future Conditioning and Latent Visualization}
\label{sec:appendix_future_masking}
The 6-frame clip forms three 2-frame temporal units, corresponding to the current state and two future targets. The model therefore predicts $K=2$ future latent states: $\hat{\mathbf{z}}_{t+1}$ for the near future and $\hat{\mathbf{z}}_{t+2}$ for the far future. Here, the prediction indices refer to temporal units rather than individual video frames. The learnable queries $\mathbf{s}_{1:K}$ are distinct from these predicted states. Near-future masking targets the predicted tokens corresponding to the first future unit, far-future masking targets those corresponding to the second, and all-future masking targets both groups.

These interventions are applied only at inference, without retraining. They differ from the Stage~III future-conditioning ablation, which changes training, and from stochastic world-model branch dropout, which omits both historical and future tokens during selected training updates.

Table~\ref{tab:future_masking} expands the inference-time masking results in Table~\ref{tab:stage_ablation} by task suite. Masking all future states lowers LIBERO-Spatial from 98.90\% to 89.60\%, while LIBERO-Goal remains at 97.40\% versus 97.10\% for the full model. The aggregate reduction therefore does not imply a benefit on every suite.
\begin{table}[H]
  \centering
  \caption{\textbf{Inference-time masking of future predictions in LIBERO.} All values are success rates (\%).}
  \label{tab:future_masking}
  \resizebox{\linewidth}{!}{
    \begin{tabular}{lccccc}
      \toprule
      Method & LIBERO-Spatial & LIBERO-Object & LIBERO-Goal & LIBERO-Long & Average \\
      \midrule
      Without all future states & 89.60 & 95.40 & 97.40 & 89.80 & 93.05 \\
      Without far-future states  & 89.60 & 97.40 & 99.80 & 95.00 & 95.45 \\
      Without near-future states  & 92.40 & 97.00 & 99.00 & 94.80 & 95.80 \\
      PLaW-VLA & 98.90 & 99.00 & 97.10 & 94.60 & 97.40 \\
      \bottomrule
    \end{tabular}
  }
\end{table}

The figure beside Table~\ref{tab:stage_ablation} provides a qualitative view of the predicted states; it does not measure prediction accuracy.

\clearpage
\section{Real-World Evaluation Protocol}
\label{sec:appendix_real_world_protocol}

\subsection{Task Specifications and Scoring}
\label{sec:appendix_task_scoring}
\begin{figure}[H]
  \centering
  \begin{subfigure}{0.8\linewidth}
    \centering
    \includegraphics[width=\linewidth]{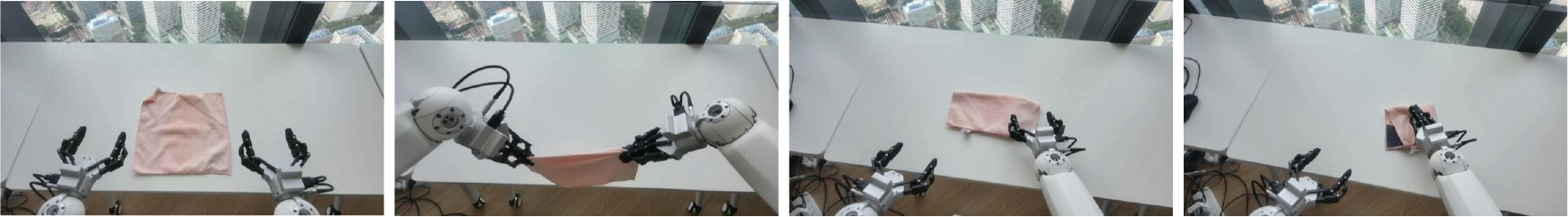}
    \caption{Fold Towel}
  \end{subfigure}
  \vspace{0.5em}
  \begin{subfigure}{0.8\linewidth}
    \centering
    \includegraphics[width=\linewidth]{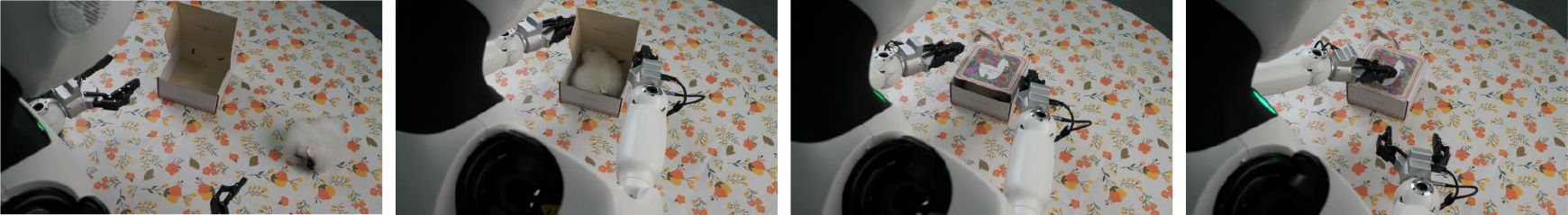}
    \caption{Pack Toy}
  \end{subfigure}
  \vspace{0.5em}
  \begin{subfigure}{0.8\linewidth}
    \centering
    \includegraphics[width=\linewidth]{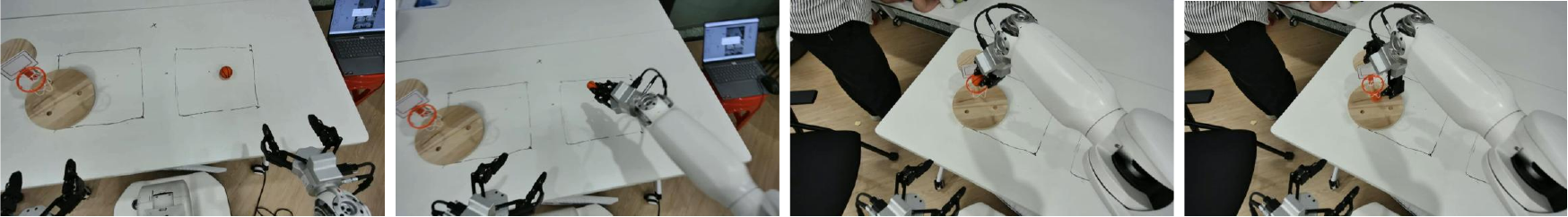}
    \caption{Play Basketball}
  \end{subfigure}
  \vspace{0.5em}
  \begin{subfigure}{0.8\linewidth}
    \centering
    \includegraphics[width=\linewidth]{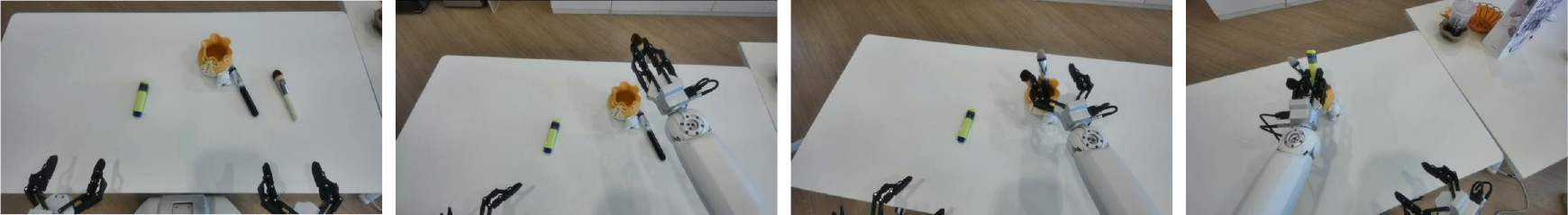}
    \caption{Organize Stationery}
  \end{subfigure}
  \caption{Visualization of real-robot fine-tuning tasks across different objects.}
  \label{fig:real_world_tasks}
\end{figure}

Fig.~\ref{fig:real_world_tasks} illustrates the four tasks. We collect 400 demonstrations per task for Stage~III adaptation, using the same task-specific training budget for PLaW-VLA, $\pi_0$, and $\pi_{0.5}$.

\paragraph{Fold Towel.}
The robot grasps the center of a deformable towel, places it on the table in a half-folded configuration, flips one edge for the second fold, and aligns the result. Scoring is binary: 1 for the correct final configuration and 0 otherwise.

\paragraph{Pack Toy.}
The robot grasps a toy, places it into a box, and adjusts the lid for proper closure. The maximum score is 3: 1 point for grasping the toy, 1 for placing it in the box, and 1 for closing the box.

\paragraph{Play Basketball.}
The robot grasps a small basketball and places it into a hoop, requiring accurate height and depth estimation. The maximum score is 2: 1 point for grasping and 1 for placement in the hoop.

\paragraph{Organize Stationery.}
The robot places three pen-like objects into a pen holder; insertion requires wrist rotation and alignment with the holder opening. Each object earns 1 point for grasping and 1 for correct placement, giving a maximum score of 6.

\subsection{Evaluation Protocol}
For each setting, we define 10 spatially distributed configurations by varying the target object or goal position across reachable regions and object--goal relations. Each method performs three trials per configuration, giving 30 trials per setting and 180 trials per method across four seen-task settings and two zero-shot variants. All methods use the same configurations. Fold Towel* changes the towel material and color; Organize Stationery* replaces the pen holder with a paper cup of different height, opening diameter, and color. The two variants are evaluated without additional adaptation. We report average task progress by normalizing each trial's score by the task maximum (1, 3, 2, or 6) and averaging over trials. This metric equals success rate for Fold Towel and credits partial completion for the other tasks.

\section{Reproducibility Details}

\subsection{Training Details}
\label{sec:appendix_training_details}

Table~\ref{tab:training_details} specifies the stage-wise batch sizes, schedules, and update budgets. Stage~III uses a linear warmup followed by a constant learning rate of $5.0\times 10^{-5}$. LIBERO warms up for 10k steps and trains for 50k steps; reported results use the 30k checkpoint. The vision-language and action experts are initialized from $\pi_{0.5}$~\cite{black2025pi_05}; the LWM expert is trained from scratch. In Stage~I, only the LWM learns from visual history and stop-gradient future targets encoded by frozen V-JEPA 2, without using action labels. Stages~II and III jointly optimize prediction and action generation using Eq.~\eqref{eq:joint_training_loss}. The action loss is always applied. With probability $\alpha=0.3$, the world-model branch is skipped, so neither its tokens nor $\mathcal{L}_{\mathrm{wm}}$ enter that update. Mixed precision and gradient accumulation are used where needed to reach the listed global batch sizes.

For the loss and conditioning ablations in Table~\ref{tab:stage_ablation}, Stages~I--II remain fixed and only Stage~III changes. Removing prediction loss retains the world-model expert and future conditioning; removing future conditioning blocks the action expert from attending to predicted states while preserving prediction loss. These variants use the same Stage~III budget and three-seed evaluation as the full model. Platform-specific action dimensions are trained jointly with arm and gripper actions during real-world adaptation.
\begin{table}[h!]
  \centering
  \caption{\textbf{Training configurations across the three stages of PLaW-VLA.}
    Stage~I corresponds to \(\mathcal{D}_{\mathrm{warm}}\), Stage~II to \(\mathcal{D}_{\mathrm{pre}}\), and Stage~III to task-specific \(\mathcal{D}_{\mathrm{post}}\).
    All stages use the same temporal sampling frequency, and all action-annotated robot data are represented in a unified end-effector (EEF) delta-action space.
    Stage~III uses a linear warmup followed by a constant learning rate. LIBERO warms up for 10k steps, trains for 50k steps, and reports the 30k checkpoint.}
  \label{tab:training_details}
  \resizebox{\linewidth}{!}{
    \begin{tabular}{l c c c c c}
      \toprule
      \multirow{2}{*}{\textbf{Hyperparameters}}
      & \multirow{2}{*}{\textbf{Stage~I}}
      & \multirow{2}{*}{\textbf{Stage~II}}
      & \multicolumn{3}{c}{\textbf{Stage~III}} \\
      \cmidrule(lr){4-6}
      & & & \textbf{LIBERO} & \textbf{RoboTwin} & \textbf{Real-world} \\
      \midrule
      Dataset Regime              & \(\mathcal{D}_{\mathrm{warm}}\)
      & \(\mathcal{D}_{\mathrm{pre}}\)
      & \(\mathcal{D}_{\mathrm{post}}\)
      & \(\mathcal{D}_{\mathrm{post}}\)
      & \(\mathcal{D}_{\mathrm{post}}\) \\
      Trainable Modules           & LWM only
      & Full MoT
      & Full MoT
      & Full MoT
      & Full MoT \\
      Batch Size                  & 512
      & 256
      & 256
      & 256
      & 256 \\
      Learning Rate               & \(5.0\times 10^{-5}\)
      & \(5.0\times 10^{-5}\)
      & \(5.0\times 10^{-5}\)
      & \(5.0\times 10^{-5}\)
      & \(5.0\times 10^{-5}\) \\
      LR Scheduler                & Constant
      & Constant
      & Warmup + constant
      & Warmup + constant
      & Warmup + constant \\
      Warmup Steps                & 0
      & 0
      & 10K
      & 10K
      & per-task (1 epoch) \\
      Training Steps              & 100K
      & 100K
      & 50K (report 30K)
      & 50K
      & per-task (\(\sim\)50 epochs) \\
      Optimizer                   & AdamW
      & AdamW
      & AdamW
      & AdamW
      & AdamW \\
      Grad. Clip Norm             & 1.0
      & 1.0
      & 1.0
      & 1.0
      & 1.0 \\
      WM Branch Dropout \(\alpha\) & --
      & 0.3
      & 0.3
      & 0.3
      & 0.3 \\
      \bottomrule
    \end{tabular}
  }
\end{table}

\clearpage
\subsection{Data Corpus and Action Alignment}
\label{sec:appendix_data_corpus}

Table~\ref{tab:datasets} identifies the data sources, frame counts, embodiments, and stage assignments needed to reproduce the training mixture. $\mathcal{D}_{\mathrm{warm}}$ supplies visual sequences for Stage~I, including action-free EgoDex videos; $\mathcal{D}_{\mathrm{pre}}$ supplies action-annotated robot trajectories for Stage~II; and $\mathcal{D}_{\mathrm{post}}$ supplies task-specific demonstrations for Stage~III. Robot action annotations are ignored in Stage~I. All datasets are resampled to 10\,Hz.
\begin{table}[h]
  \centering
  \caption{\textbf{Heterogeneous data corpus used for Stage~I and Stage~II.}
    Stage~I uses visual sequences from all five sources without action annotations. Stage~II uses the four robot datasets with actions aligned to a shared end-effector representation. EgoDex is used only in Stage~I. Stage~III settings appear in Table~\ref{tab:training_details}.}
  \label{tab:datasets}
  \resizebox{\linewidth}{!}{
    \begin{tabular}{l l c c c}
      \toprule
      Data Source & Embodiment & Num. Frames & Stage~I & Stage~II \\
      \midrule
      InternData-A1~\cite{tian2025interndata}  & Cross-embodiment (Genie-1, Franka Panda, AgileX ALOHA, ARX Lift-2) & 377.7M & \checkmark & \checkmark \\
      AgiBotWorld~\cite{bu2025agibot_iros}     & Bimanual Mobile (AgiBot G1)                             & 56.9M  & \checkmark & \checkmark \\
      RoboTwin 2.0~\cite{chen2025robotwin}            & Bimanual (AgileX ALOHA)  & 6.1M   & \checkmark & \checkmark \\
      LIBERO-90~\cite{liu2023libero}              & Franka Panda                                                & 0.6M   & \checkmark & \checkmark \\
      EgoDex~\cite{egodex}                        & Human (egocentric, hand-tracking)                     & 77.1M  & \checkmark & --         \\
      \bottomrule
    \end{tabular}
  }
\end{table}

\paragraph{Cross-embodiment action alignment.}
We map action-annotated robot trajectories into a shared EEF representation. Each arm uses an 8-dimensional action
\[
  a=[\Delta x,\Delta y,\Delta z,q_w,q_x,q_y,q_z,g],
\]
where $(\Delta x,\Delta y,\Delta z)$ is relative EEF translation, $(q_w,q_x,q_y,q_z)$ is the normalized EEF orientation quaternion, and $g$ is the gripper command. Bimanual actions concatenate the two arms into a 16-dimensional vector. Stage~III adds platform-specific dimensions, such as torso, head, or mobile-base control, when required by the target interface. Where applicable, these additional components use delta control.

For LIBERO Stage~III, each absolute action target in the predicted action chunk is converted into a delta relative to the current robot state $s_t$. Translation and gripper targets are offset from the current state, while orientation is represented as a normalized relative quaternion. All future actions in the chunk use the same current-state reference,
\begin{equation}
  a_{t+i}=\Delta(s_t,\tilde a_{t+i}), \qquad i=0,\ldots,H_a-1,
\end{equation}
where $\tilde a_{t+i}$ is the absolute target at chunk index $i$ and $H_a$ is the chunk length.

\end{document}